%% file: submission.tex
\documentclass[letterpaper, 10 pt, conference]{ieeeconf}  

\IEEEoverridecommandlockouts                              

\usepackage{amsmath} 
\usepackage{amssymb}  
\usepackage{cite}
\usepackage{booktabs}
\usepackage[subnum]{cases}
\usepackage{mathtools}
\usepackage{balance}
\usepackage{color}
\usepackage{tabularx}
\usepackage{graphicx,dblfloatfix} 
\usepackage{listings}
\usepackage{textcomp}
\usepackage{url}
\usepackage{multirow}
\usepackage{algorithm}
\usepackage{algorithmic}
\usepackage{ragged2e} 
\newcolumntype{Y}{>{\RaggedRight\arraybackslash}X} 

\usepackage{bm}
\usepackage{bbm}
\usepackage{hyperref}
\usepackage[normalem]{ulem}
\usepackage{units}
\usepackage{threeparttable} 
\usepackage{pifont}
\usepackage[T1]{fontenc}
\usepackage{eucal}

\usepackage[printonlyused]{acronym}
\input{acronyms} 

 \newcommand{\deleted}[1]{}

\title{\LARGE \bf
Advanced Skills through Multiple Adversarial Motion Priors in Reinforcement Learning
}

\author{Eric Vollenweider, Marko Bjelonic, Victor Klemm, Nikita Rudin, Joonho Lee and Marco Hutter
\thanks{
This work was supported in part by armasuisse W\&T and the Swiss National Science Foundation (SNF) through the National Centres of Competence in Research Robotics (NCCR Robotics) and Digital Fabrication (NCCR dfab). Besides, it has been conducted as part of ANYmal Research, a community to advance legged robotics.}
\thanks{All authors are with the Robotic Systems Lab, ETH Z\"urich, 8092 Z\"urich, Switzerland.}
\thanks{{\tt\footnotesize ericvol@microsoft.com},
{\tt\footnotesize marko.bjelonic@mavt.ethz.ch},
{\tt\footnotesize victor.klemm@mavt.ethz.ch},
{\tt\footnotesize nikita.rudin@mavt.ethz.ch},
{\tt\footnotesize joonho.lee@mavt.ethz.ch}
}
}

\begin{document}


\maketitle

\begin{abstract}
\input{0_abstract}
\end{abstract}



\input{1_introduction}
\input{2_method}

\input{3_experimental_results}
\input{4_conclusion}







\balance
\bibliographystyle{IEEEtran}
\bibliography{IEEEabrv,submissionbibfile}

\end{document}

%% file: acronyms.tex
\newcommand{\acr}[1]{\acs{#1} (\aclu{#1})}
\acrodef{RSL}{Robotic Systems Lab}
\acrodef{COM}{center of mass}
\acrodef{SQP}{sequential quadratic problem}
\acrodef{VMC}{virtual model controller}
\acrodef{w.r.t.}{with respect to}
\acrodef{DOF}{degrees of freedom}
\acrodef{ZMP}{zero moment point}
\acrodef{COP}{center of pressure}
\acrodef{IMU}{inertial measurement unit}
\acrodef{COT}{cost of transport}
\acrodef{JPL}{Jet Propulsion Laboratory}
\acrodef{HAA}{hip adduction/abduction}
\acrodef{HFE}{hip flexion/extension}
\acrodef{KFE}{knee flexion/extension}
\acrodef{ZMP}{zero-moment point}
\acrodef{QP}{quadratic programming}
\acrodef{SQP}{sequential quadratic programming}
\acrodef{WBC}{whole-body controller}
\acrodef{HO}{hierarchical optimization}
\acrodef{NLP}{nonlinear programming}
\acrodef{MPC}{model predictive control}
\acrodef{TO}{trajectory optimization}
\acrodef{DARPA}{Defense Advanced Research Projects Agency}
\acrodef{JPL}{Jet Propulsion Laboratory}
\acrodef{RBDL}{Rigid Body Dynamics Library}
\acrodef{RL}{reinforcement learning}
\acrodef{LIP}{linear inverted pendulum}
\acrodef{COP}{center of pressure}
\acrodef{SLQ}{sequential linear quadratic}
\acrodef{DDP}{differential dynamic programming}
\acrodef{SRBD}{single rigid body dynamics}
\acrodef{EOM}{equations of motion}
\acrodef{AOW}{ANYmal on Wheels}
\acrodef{RL}{Reinforcement Learning}
\acrodef{AMP}{Adversarial Motion Priors}
\acrodef{PPO}{Proximal Policy Optimization}
\acrodef{TRPO}{Trust Region Policy Optimization}
\acrodef{SAC}{Soft Actor Critic}
\acrodef{GAN}{Generative Adversarial Network}

%% file: 0_abstract.tex
In recent years, reinforcement learning (RL) has shown outstanding performance for locomotion control of highly articulated robotic systems.
Such approaches typically involve tedious reward function tuning to achieve the desired motion style. Imitation learning approaches such as adversarial motion priors aim to reduce this problem by encouraging a pre-defined motion style.
In this work, we present an approach to augment the concept of adversarial motion prior-based RL to allow for multiple, discretely switchable styles.
We show that multiple styles and skills can be learned simultaneously without notable performance differences, even in combination with motion data-free skills.
Our approach is validated in several real-world experiments with a wheeled-legged quadruped robot showing skills learned from existing RL controllers and trajectory optimization, such as ducking and walking, and novel skills such as switching between a quadrupedal and humanoid configuration. For the latter skill, the robot is required to stand up, navigate on two wheels, and sit down. Instead of tuning the sit-down motion, we verify that a reverse playback of the stand-up movement helps the robot discover feasible sit-down behaviors and avoids tedious reward function tuning.

%% file: 1_introduction.tex
\section{INTRODUCTION}
\begin{figure}[t!]
    \centering
    \includegraphics[width=0.95\columnwidth]{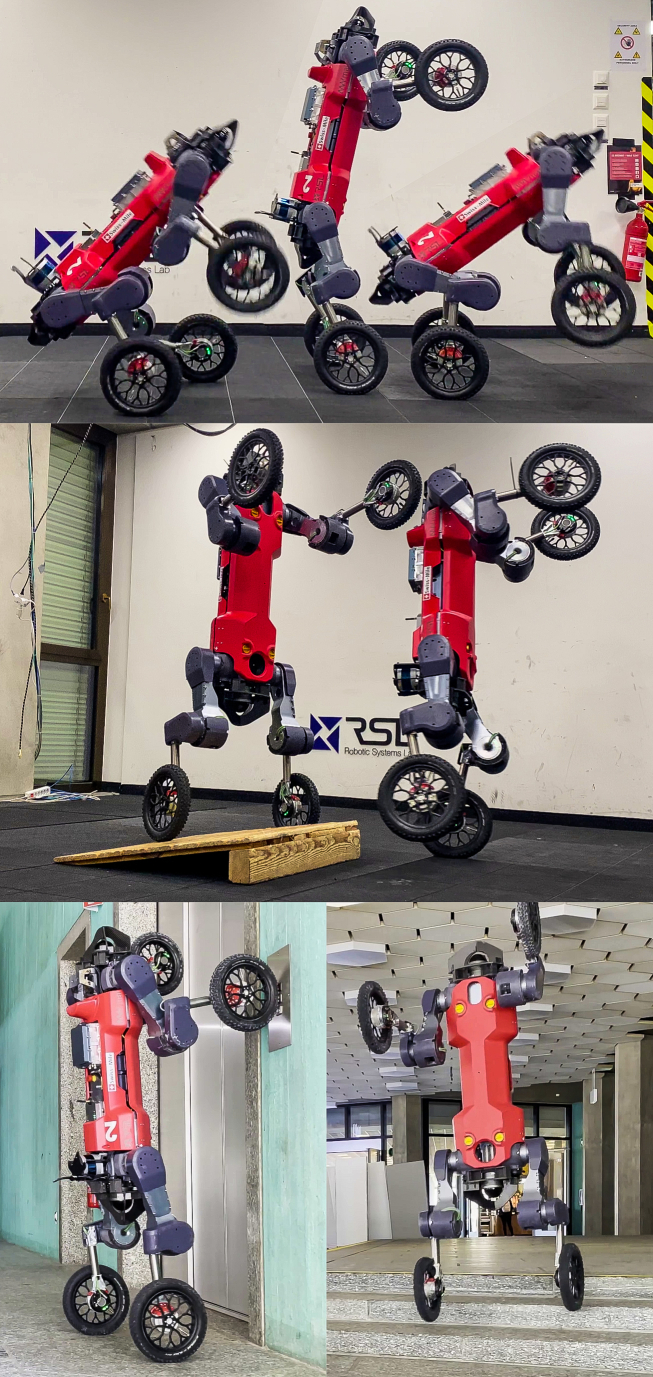}
    \caption{Quadruped-humanoid transformer (\href{ https://youtu.be/kEdr0ARq48A}{https://youtu.be/kEdr0ARq48A}) with a time-lapse from left to right of a stand-up and sit-down motion (top image), obstacle negotiation (middle image), and indoor navigation (bottom images). The former skill and the humanoid navigation on two legs are achieved through traditional RL training with a task reward formulation. Instead of tuning the sit-down skill, we can reverse the playback of the stand-up motion and use it as a motion prior that helps the robot discover feasible sit-down behaviors avoiding tedious reward function tuning.}
    \label{fig:first_page}
\end{figure}

\ac{RL} had a significant impact in the space of legged locomotion, showcasing robust policies that can handle a wide variety of challenging terrain in the real world~\cite{miki2022learning}. With this advancement, we believe that these articulated robots can perform specialized motions like their natural counterparts. Therefore, we aim to push these robots even more to their limits by executing advanced skills like the \emph{quadruped-humanoid transformer} in Fig.~\ref{fig:first_page} performed by our wheeled-legged robot~\cite{bjelonic2021wholebody}. In this work, we rely on a combination of motion priors and \ac{RL} to achieve such skills.

\subsection{Related Work}
Executing specific behaviors for a real robot is a fundamental challenge in robotics and \ac{RL}. For example, the computer animation community synthesizes life-like behaviors from human or animal demonstrations for their simulated agents. Boston Dynamic's real humanoid robot, \emph{Atlas}, shows impressive dancing motions and backflips based on human motion animators. Similarly, our wheeled-legged robot can track motions from an offline trajectory optimization with an \ac{MPC} algorithm, as shown in our previous work~\cite{bjelonic2022complex}. Furthermore, motion optimizations, such as \cite{physicsbased1, pysicsbased2}, have the added benefit of producing physically plausible motions, which is favorable in computer graphics but vital in robot control. However, designing objective functions is usually exceptionally difficult. However, these tracking-based methods require carefully designed objective functions. When applied to more extensive and diverse motion libraries, these methods need heuristics to select the suitable motion prior to the scenario.

Data-driven strategies like \cite{peng2021amp} automate the imitation objective and mechanisms for motion selection based on adversarial imitation learning. This paper verifies that this imitation learning approach can be applied to real robotics systems and not just computer animations. Gaussian processes~\cite{LevineGaussians, Guassian2} can learn a low-dimensional motion embedding space generating suitable kinematic motions when provided with a relatively large amount of motion data. However, the approaches are not goal conditioned and can not leverage task-specific information. 

Animation techniques \cite{imitation1,imitation2RL,2018-TOG-deepMimic} attempt to solve this by imitating/tracking motion clips. This is usually implemented with pose errors, requiring a motion clip selection and synchronizing the selected reference motion and the policy's movement. By using a phase variable as an additional input to the policy, the right frame in the motion data-set can be selected. It can be challenging to scale the number of motion clips with these approaches. Defining error metrics that generalize to a wide variety of motions is difficult.

Two alternative approaches are adversarial learning and student-teacher architectures~\cite{LearningLocoANYmalStudentTeacher}. The latter trains a teacher policy with privileged information such as perfect knowledge about the height map, friction coefficients, and ground contact forces. With that, the teacher can learn complex motions more easily. After the teacher's training, the student policy learns to reproduce the teacher's output using non-privileged observations and the robot's proprioceptive history. Hereby, a style transfer from teacher to student is happening. On the other hand, adversarial imitation learning techniques \cite{AdvImit1, AdvImit2} and more recently \cite{peng_ma_abbeel_levine_kanazawa_2021} build upon a different approach. The latter offers a discriminator-based learning strategy called \ac{AMP}, which outsources the error-metrics, phase, and motion-clip selection to a discriminator which learns to distinguish between the policy's and motion data's state transitions. \emph{\ac{AMP}} does not require specific motion clips to be selected as tracking targets since the policy automatically chooses which style to apply given a particular task. The method's limitation is that whenever multiple provided motion-priors cover the same task, the policy might either go for the more straightforward style to fulfill or find a hybrid motion similar to both motion clips. In other words, there is no option of actively choosing styles in single or multi-task settings. Furthermore, the task-reward still has to motivate the policy to execute a specific movement because otherwise the policy might identify two states and oscillate between them. Generally, to our experience, it is not trivial to find task-reward formulations for complex and highly dynamic movements that do not conflict with the style reward provided by the discriminator. 
\subsection{Contribution}
This paper introduces the Multi-\ac{AMP} algorithm and applies it to our real wheeled-legged robot. Like its \ac{AMP} predecessor~\cite{peng2021amp}, this approach automates the imitation objective and motion selection process without heuristics. Furthermore, our extension allows for the intentional switching of multiple different style objectives. The approach can imitate motion priors from three different data sets, i.e., from existing \ac{RL} controllers, trajectory optimization, and reverse stand-up motions. The latter enables the automatic discovery of feasible sit-down motions on the real robot without tedious reward function tuning. This permits exceptional skills with our wheeled-legged robot in Fig.~\ref{fig:first_page}, where the robot can switch between a quadruped and humanoid configuration. To the best of our knowledge, this is the first time such a highly dynamic skill is shown and also the first time that the \ac{AMP} approach is verified on a real robot.

%% file: 2_method.tex
\section{MULTIPLE ADVERSARIAL MOTION PRIORS}

\newcommand{\mmpriori}{M^{i}}
\newcommand{\mpriori}{$\mmpriori$~}
\newcommand{\mmpolicy}{B^i_\pi}
\newcommand{\mpolicy}{$\mmpolicy$~}
\newcommand{\rtaskt}{$r_t^{task}$} 

In this work, the goal is to train a policy $\pi$ capable of executing multiple tasks, including styles extracted from $n$ individual motion data-sets $M^i, i \in \{0, ..., n-1\}$ with the ability to actively switch between them. In contrast to tracking-based methods, the policy should not blindly follow specific motions but rather extract and apply the underlying characteristics of the movements while fulfilling its task.

\begin{figure}
    \centering
    \includegraphics[width=0.48\textwidth]{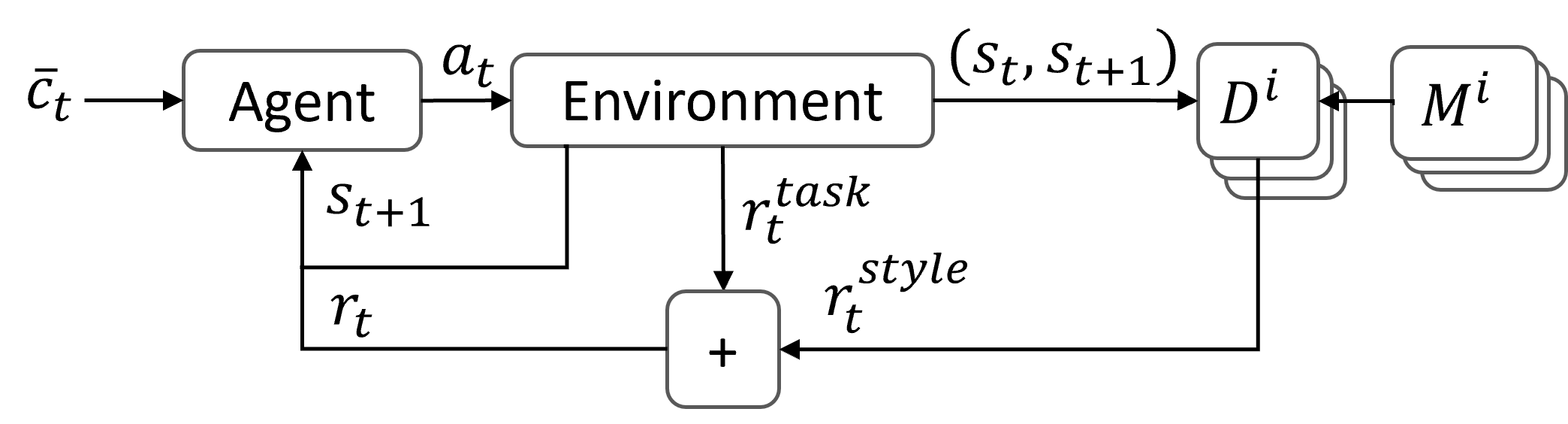}
    \caption{Multi-AMP overview: The discriminator predicts a style reward $s_t^{style}$ which is high if the policy's behavior is similar to the motions of the motion-data base $M^i$, by distinguishing between state transitions $(s_t, s_{t+1})$ of both sources. The style reward is added to the task reward, which finally leads to the policy fulfilling the task while applying the motion data's style.}
    \label{fig:architecture}
\end{figure}

Similar to the \ac{AMP} algorithm~\cite{peng2021amp}, we split the reward calculation into two parts $r_t = r_t^{task} + r_t^{style}$. The task-reward is a description of \emph{what} to do, e.g., velocity tracking, and the style-reward $r_t^{style}$ defines \textit{how} to do it, namely by extracting and applying the style of the motion priors. While task rewards often have simple mathematical descriptions, the style reward is not trivial to calculate. In the following, we introduce \emph{Multi-\ac{AMP}}, a generalization of \emph{AMP} which allows for switching of multiple different style-rewards, which constitutes the main theoretical contribution of this work. 

A style reward motivates the agent to extract the motion prior's style. We use an adversarial setup with $n$ discriminators $D^i, i \in \{0, ..., n-1\}$. For every trained style $i$, a roll-out buffer \mpolicy collects the states of time-steps where the policy applies the $i^{th}$ style, and another buffer \mpriori contains the motion-data prior to that specific style. Each discriminator $D^i$ learns to differentiate between descriptors built from a pair of consecutive states $(s_t, s_{t+1})$ sampled from \mpriori and \mpolicy. Thus, every trainable style is defined by a tuple $\{D^i, \mmpolicy, \mmpriori\}$. By avoiding any dependency on the source's actions, the pipeline can process data of sources with unknown actions, such as data from motion-tracking and character animation. The discriminator $D^i$ learns to predict the difference between random samples of its motion database \mpriori, and the agent's transitions sampled from the style's roll-out buffer \mpolicy by scoring them with $+1$ and $-1$, respectively.
This behavior is encouraged by solving the least-squares problem \cite{peng2021amp} defined by
\begin{equation} \label{eq:multi_amp_disc_loss}
\begin{split}
    L^i =~&\mathbb{E}_{d^{\mmpriori}(s,s')} \left [ (D^i(\phi(s), \phi(s')) - 1)^2 \right] \\
      +~&\mathbb{E}_{d^{\mmpolicy(s,s')}}~\left [ (D^i(\phi(s), \phi(s')) + 1)^2 \right] \\
      +~&\frac{w^{gp}}{2} \mathbb{E}_{d^{\mmpriori}(s,s')} \left[ \|  \nabla_{\phi D^i(\phi)}|_{\phi = (\phi(s), \phi(s'))}  \|^2 \right],
\end{split}
\end{equation}

where the descriptors are built by concatenating the output of an arbitrary function $\phi(\cdot): \mathbb{R}^{d_s} \mapsto \mathbb{R}^{d_d}$ for two consecutive states, whereby the choice of $\phi$ decides which style information is extracted from the state-transitions, e.g., the robot's joint and torso position, velocity, etc.

\subsection{Style-reward}
During the policy's roll-out only one style is active at a time. The state $s_t$ passed into the policy at every time-step $t$ contains a command $c_t$, which is augmented with a one-hot-encoded style selector $c_{s}$, i.e., the elements of  $c_{s}$ are zero everywhere except at the index of the active style $i$. As in the standard RL-cycle, after the policy $\pi(a_t|s_t)$ predicts an action $a_t$, the environment returns a new state $s_{t+1}$ and a task-reward $r_t^{task}$. The latest state-transition $(s_t, s_{t+1})$ is used to construct the style-descriptor $d_t = [\phi(s_t), \phi(s_{t+1})] \in \mathbb{R}^{2d_d}$, which is mapped to a style-reward $r_t^{style} \in \mathbb{R}^+$ using the current style's discriminator $D^i$ and the style-reward given by
\begin{equation}
\label{eq:amp_reward}
    r^{style}_t = -log\left(1 - \frac{1}{1 + \exp^{-D^i([\phi(s_t), \phi(s_{t+1})])}} \right).
\end{equation}
\subsection{Task-reward}
Our agents interact with the environment in a command-conditioned framework. During the training, the environment rewards the policy for fulfilling commands $c_t$ sampled from a command distribution $p(c)$. For example, the task might be to achieve the desired body velocity sampled from a uniform distribution in x, y and yaw coordinates. The task is included in the policy's observation and essentially informs the agent what to do. The task reward depends on the performance of the policy with respect to the command $r_t^{task} = R(c_t, s_t, s_{t-1})$
\subsection{Multi-AMP algorithm}
The sum of the style and task rewards $r_t = r_t^{task} + r_t^{style}$ constitutes the overall reward, which can be used in any \ac{RL} algorithm such as \ac{PPO}~\cite{schulman2017proximal} or \ac{SAC}~\cite{haarnoja2018soft}. The state $s_t$ is additionally stored in the style's roll-out buffer \mpolicy to train the discriminator at the end of the epoch. The full approach is shown in the following algorithm:

\begin{algorithmic}[1]
\label{alg:multi_amp}
\REQUIRE $M = \{M_i\}, |M| = n$ ~~ (n motion data-sets)
\STATE $\pi \gets $ initialize policy
\STATE $V \gets $ initialize Value function
\STATE $[\mathcal{B}] \gets$ initialize n style replay buffers
\STATE $[D] \gets $ initialize $n$ discriminators
\STATE $\mathcal{R} \gets$ initialize main replay buffers

\WHILE{not done}
    \FOR{trajectory i = 1, ..., m}
        \STATE $\tau^i \gets \{ (c_{t}, c_{s}, s_t, a_t, r_t^G)_{t=0}^{T-1}, s_T, g\}$ roll-out with $\pi$
        
        \STATE $d \gets$ style-index of $\tau^i$ (encoded in $c_{s}$)
        
        \IF {$M^d$ is not empty} \label{alg:line:empty_dataset}
            \FOR{t = 0, ..., T-1}
                \STATE $d_t \gets D^d(\phi(s_t), \phi(s_{t+1}))$
                \STATE $r_t^{style} \gets $ according to Eq. \ref{eq:amp_reward}
                \STATE record $r_t^{style}$ in $\tau^i$
            \ENDFOR
            \STATE store $d_t$ in $\mathcal{B}^d$ and $\tau_i$ in $\mathcal{R}$
        \ENDIF
            
    \ENDFOR
    
    \FOR{update step = 1, ..., $n_{updates}$}
        \FOR{d = 0, ..., n}
            \STATE $b^\mathcal{M} \gets$ sample batch of $K$ transitions $\{s_j, s_j'\}_{j=1}^K$ from $\mathcal{M}^d$
            \STATE $b^\pi \gets$ sample batch of $K$ transitions $\{s_j, s_j'\}_{j=1}^K$ from $\mathcal{B}^d$
            \STATE update $D^d$ according to Eq. \ref{eq:multi_amp_disc_loss}
        \ENDFOR
    \ENDFOR
    
    \STATE update $V$ and $\pi$ (standard PPO step using $\mathcal{R}$)
    
\ENDWHILE
\end{algorithmic}

\subsection{Data-free skills}
If no motion data is present for the desired skill and it should nevertheless be trained alongside multiple motion-data skills, Multi-AMP can be adapted slightly. While the policy learns the motion-data free skill, $r_t^{style}$ is set to $0$. Thereby, the data-free skill is still treated as a valid style and present in the one-hot-encoded style-selector $c_s$, but the policy $\pi$ is not guided by the style-reward anymore.

%% file: 3_experimental_results.tex
\section{EXPERIMENTAL RESULTS AND DISCUSSION}
\label{sec:experiments}
\begin{figure*}
\centering
\includegraphics[width=\linewidth]{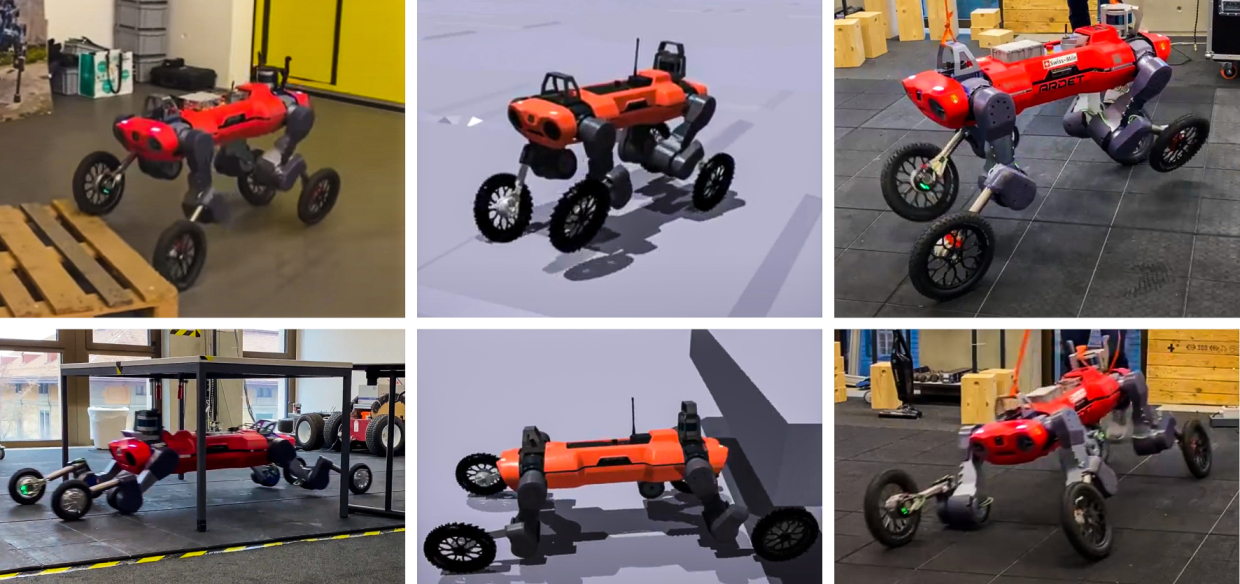}

\caption{Four-legged locomotion (top row) and ducking motion (bottom row) of the motion data source (left column), simulation training (center column), and final deployment on the real robot using Multi-AMP. The former skill is trained with a motion prior from a different simulation environment and control approach, while the ducking motion is trained with data from trajectory optimization~\cite{bjelonic2022complex}.}
\label{fig:3styletasks}
\end{figure*}

\begin{figure*} \centering
\includegraphics[width=\linewidth]{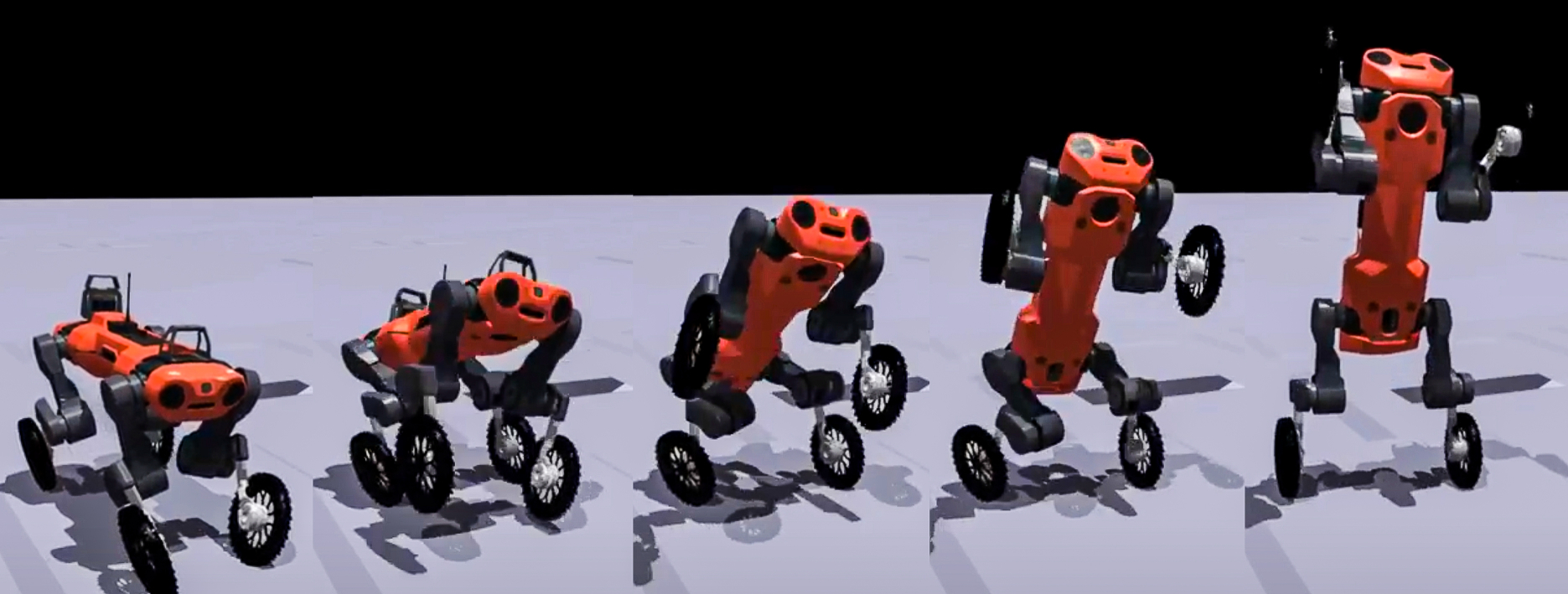}
\includegraphics[width=\linewidth]{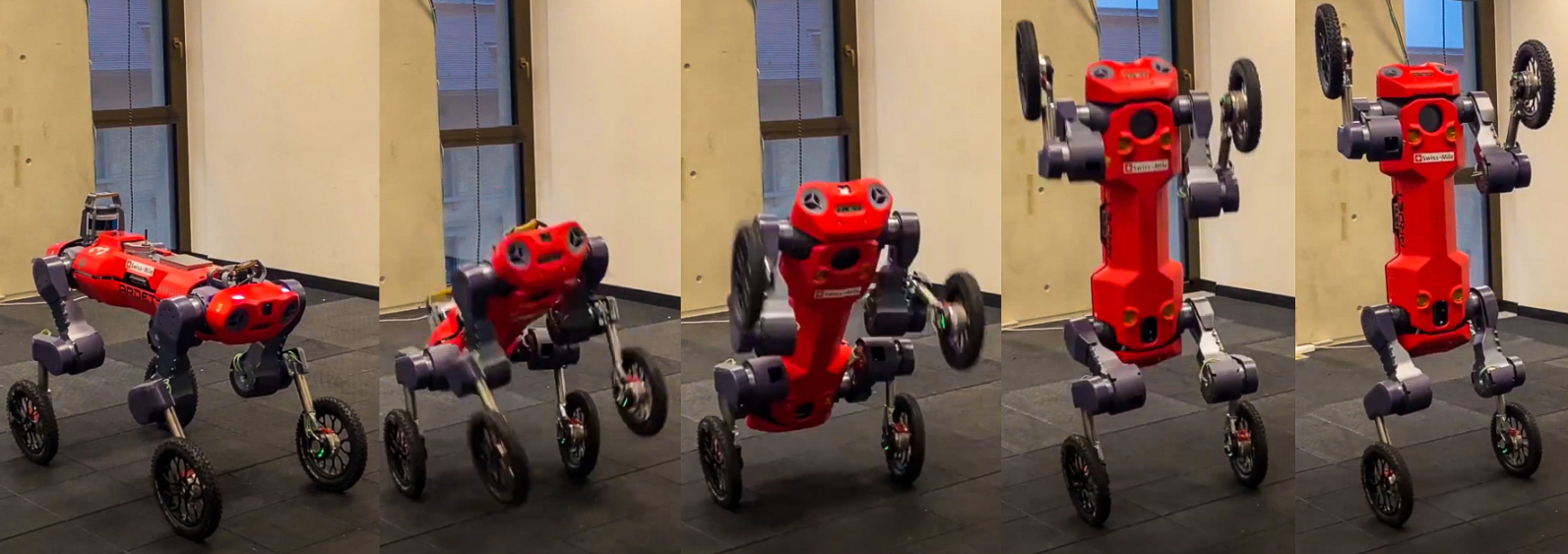}
\caption{Stand up-sequence in simulation and on the real robot. The policy is able to stand up, navigate large distances on two legs, and finally sit down again using the stand-up motion prior.}
\label{fig:standup_sequence}
\end{figure*}

We implement and deploy the proposed Multi-AMP framework on our wheeled-legged robot in Fig.~\ref{fig:first_page} with 16 \acr{DOF}. The training environment consists of three tasks, two of which are supported by motion data, and one is a data-decoupled task. The first task is four-legged locomotion, the motion data of which consists of motions recorded from another RL policy (Fig.~\ref{fig:3styletasks} top left). The second task is a ducking skill, allowing the robot to duck under a table. The motion data for this skill was generated by a trajectory optimization pipeline, which was deployed and tracked by an MPC controller~\cite{bjelonic2022complex} (Fig.~\ref{fig:3styletasks} bottom left). The last skill represents a partly data-decoupled skill. Here, the wheeled-legged robot learns to stand up on its hind legs followed by two-legged navigation (Fig.~\ref{fig:standup_sequence}), before sitting down again. The sit-down skill is supported by motion data as detailed in Section~\ref{sec:sitdown}. A video available at \href{https://youtu.be/kEdr0ARq48A}{https://youtu.be/kEdr0ARq48A} showing the results accompanies this paper.

The training environment of our Multi-AMP pipelines is implemented using the \emph{Isaac Gym simulator}~\cite{isaacgym2021,rudin2021learning}, which allows for massively parallel simulation. We spawn 4096 environments in parallel to learn all three tasks simultaneously in a single neural network. The number of environments per task is weighted according to their approximate difficulty, e.g., $[1,1,5]$ in the case of the tasks described above. The state-transitions collected during the roll-outs of these environments are mapped using a function $\phi(s)$ such that it extracts the linear and angular base velocity, gravity direction in base frame, the base's height above ground, joint position and velocity, and finally the position of the wheels relative to the robot's base-frame, i.e., $\phi(s) = (\dot{x}_{base}, x_z, e_{base}, q, \dot{q}, x_{ee, base}) \in \mathbb{R}^{50}$. The task reward definitions for the three tasks are in Table~\ref{table:task_rewards} and~\ref{table:standup_reward}.

\input{9_RewardTables}

\newcommand{\customFit}{\hspace{0.01\textwidth}}

\subsection{Experiments}
Due to the problem of catastrophic forgetting \cite{deepRLcatastrophicForgetting1, deepRLcatastrophicForgetting2, deepRLcatastrophicForgetting3}, we learn these skills in parallel. This section analyzes the task performance of each Multi-AMP policy compared to policies that exclusively learn a single task (baseline). The three tasks (standing up, ducking, and four-legged locomotion) are trained in different combinations, where ducking and walking are always learned with motion data and stand-up without:
\begin{enumerate}
    \item Stand up only
    \item Duck only
    \item Walk only
    \item Walking and standing up
    \item Walking and ducking
    \item Walking, ducking, and standing up
\end{enumerate}

\begin{figure}
    \centering
    \includegraphics[width=0.485\textwidth]{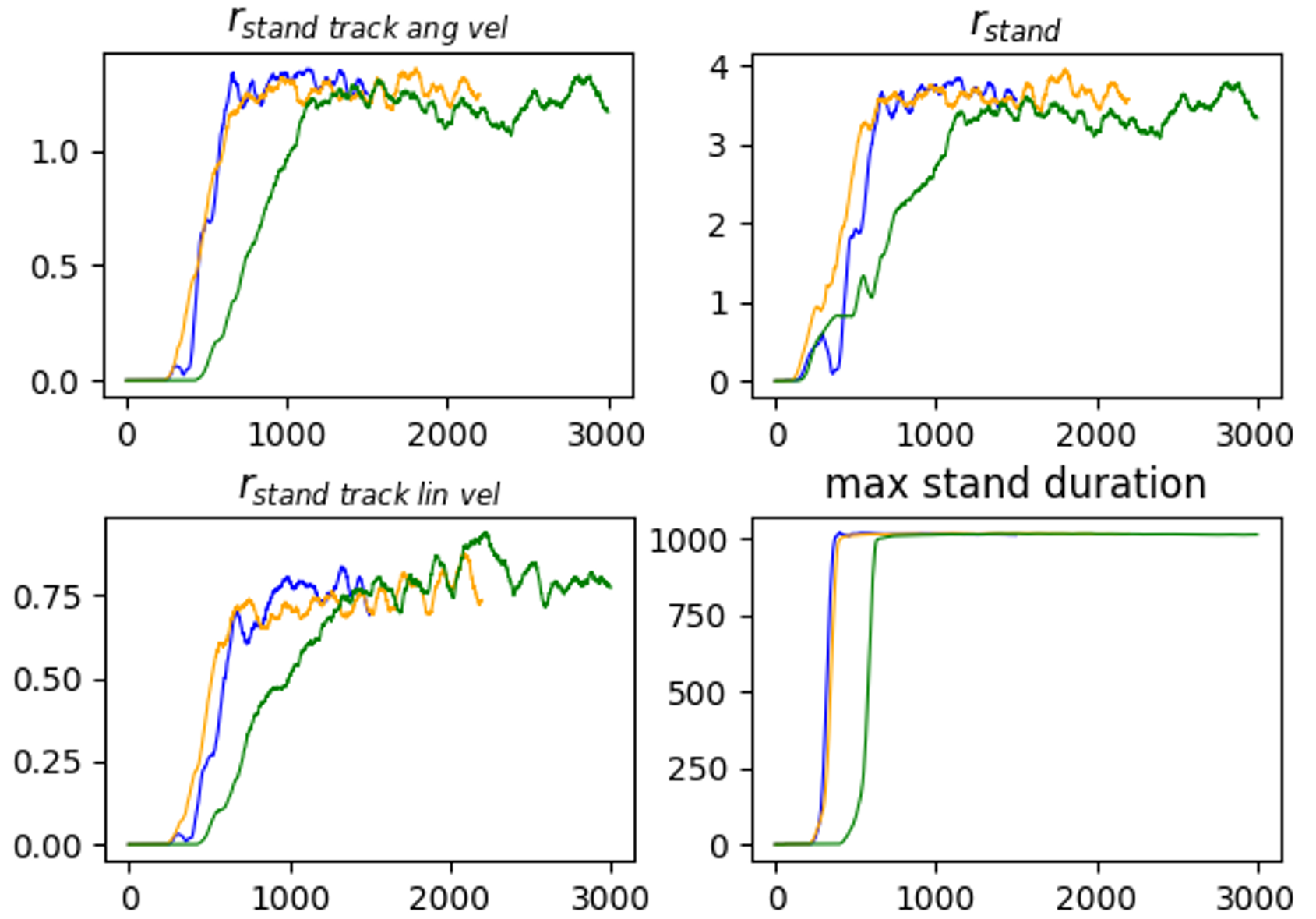}
    \caption{Multi-AMP learning capability of the stand-up task. The horizontal axis denotes the number of epochs, and the vertical axis represents the value of the reward calculations after post-processing for comparability. Furthermore, the maximum stand duration is plotted over the number of epochs. Legend: Blue (one style), yellow (two styles), blue (three styles)}
    \label{fig:multi_style_test_stand}
\end{figure}

First, we compare the learning performance of the stand-up skill between the models Nr.~1, 4, and 6. The stand-up task is an informative benchmark since it requires a complex sequence of movements to achieve the goal. We normalize all rewards in the following Figures with the number of robots receiving the reward, making the plots comparable between the experiments. Fig.~\ref{fig:multi_style_test_stand} shows important metrics of the stand-up learning progress. The figure shows that the policy does not make compromises during the training of multiple tasks compared to single-task settings. The policy that learns three tasks simultaneously (\textit{3 styles} in Fig.~\ref{fig:multi_style_test_stand}) performs equally well while standing up and sitting down. While it takes the \textit{3 style} policy a bit longer to reach the maximum rewards (see $r_{stand}$ and $r_{stand~track~ang~vel}$ at epoch 1000), the differences vanish after sufficiently long training times. In this case, it takes Multi-AMP about 300 epochs longer to reach the maximum task rewards compared to the single task policy. 

The walking and ducking tasks  show a very similar picture, with the specialized policies (model Nr. 2 and 3 in the list above) reaching a similar final performance compared to the others. Furthermore, all policies manage to extract the walking and ducking style such that no visible difference can be seen.

In summary, in this specific implementation of the environment and selection of tasks, Multi-AMP, while taking longer, learns to achieve all goals equally well as more specialized policies that learn fewer tasks.

\begin{figure}
    \centering
    \includegraphics[width=\columnwidth]{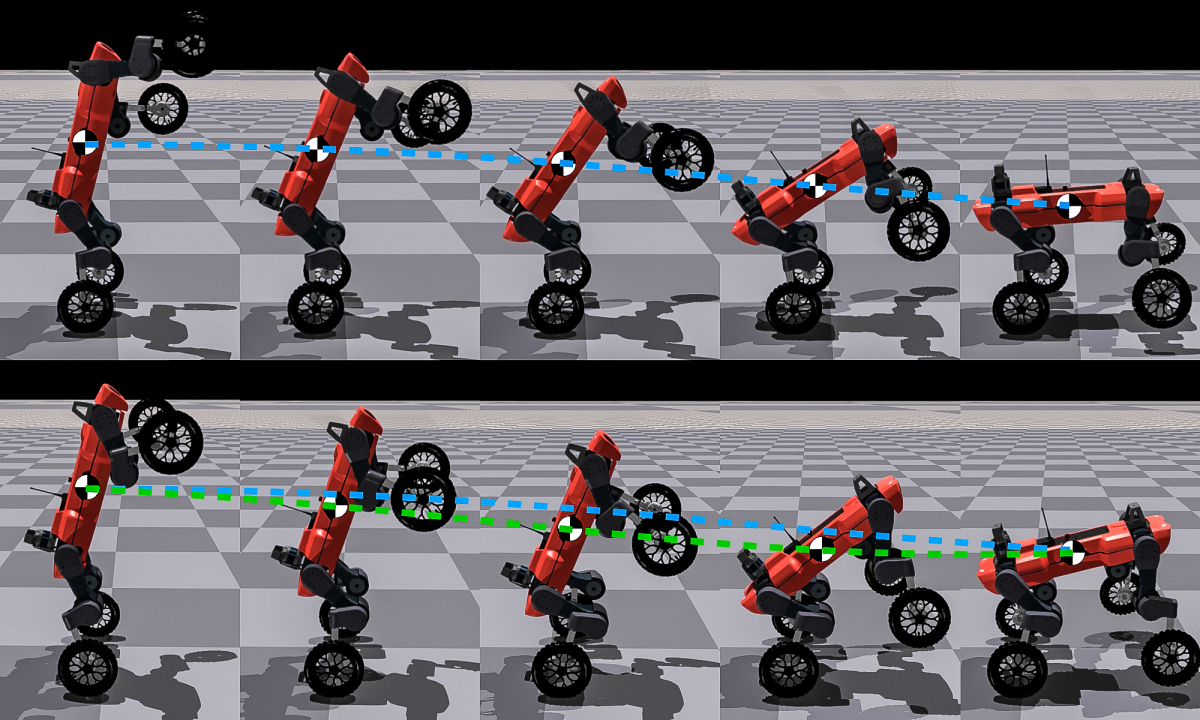}
    \caption{Comparison of the sitting down motions. 
    Top row: If the agent learns to sit down with task rewards only, it falls forward with extended front legs, which causes high impacts and leads to over-torque on the real robot. Marked in blue is the trajectory of the center of gravity of the base.
    Bottom row: When sitting down with task reward and style reward from the reversed stand-up sequence, the robot squats down to lower its center of gravity before tilting forward, thereby reducing the impact's magnitude. Marked in green is the trajectory of the center of gravity of the base. We note that compared to the previous case the base is lowered in a way that causes less vertical base velocity at the moment of impact.}
    \label{fig:sitdown_no_amp_vs_amp}
\end{figure}

\subsection{Sit-down training}\label{sec:sitdown}
While the sit-down rewards presented in Table~\ref{table:standup_reward} work well in simulation, the policy's sit-down motions created high impulses in the real robot's knees, which exceeded the robot's safety torque threshold. To easily perform more gentle sit-down motions and avoid reward function tuning, we recorded the stand-up motion, reversed the motion data, and trained a policy using Multi-AMP. As this motion starts with a front end-effector velocity of 0 when lifting them off the ground, the reversed style should encourage low impact sit-down motions. 
In the Multi-AMP combination, one style contains the reversed motion data for sitting down, while the second style receives plain stand-up rewards. The result is a sit-down motion that uses its hind knees to lower the center of gravity before tilting the base and catching itself on four legs, as shown in Fig.~\ref{fig:sitdown_no_amp_vs_amp}. The agent receives zero task rewards for a predefined time after the command to sit down, avoiding task rewards that conflict with the sit-down motion-prior. E.g., rewarding horizontal body orientation leads the agent to accelerate the sit-down, which breaks the style. After this buffer-time, the sit-down task-rewards become active and reward the agent. This allows the robot to sit down with its own speed and style and guarantees non-conflicting rewards.

\subsection{Remarks}
Finding a balance between training the policy and the discriminators is vital during the Multi-AMP training process. Our observations show that fast or slow training of the discriminators relative to the policy hampers the policy's style training. In our current implementation, the number of discriminator and policy updates is fixed, which might not be an optimal strategy. Since the setup is very similar to \ac{GAN}, more ideas from \cite{SalimansGZCRC16GAN} could be incorporated into Multi-AMP.

We use an actuator model for the leg joints to bridge the sim-to-real gap \cite{sea_actuator_model} while an actuator model is not needed for the velocity controlled wheels. Moreover, we apply strategies to increase the policy's robustness, such as rough terrain training (see rough terrain robustness in Fig.~\ref{fig:first_page}), random disturbances, and game inspired curriculum training~\cite{rudin2021learning}. The highly dynamic stand-up skill is especially prone to these robustness measures, which we solve by introducing timed pushes and joint-velocity-based trajectory termination. The former identifies the most critical phase of the skill and pushes the policy in the worst possible way. This increases the number of disturbances the policy experiences during these critical phases, rendering it more robust, and thus, also helping with sim-to-real efforts. Furthermore, by terminating the trajectory if the joint velocity of any \ac{DOF} exceeds the actuator's limits, the policy learns to keep a safety tolerance to these limits.

%% file: 9_RewardTables.tex
\begin{table}
    \caption{Task-rewards.}
    \label{table:task_rewards}
    \begin{center}
    \begin{tabular}{@{}l|ll@{}}\toprule
        All tasks & formula & weight \\
        $r_{\tau}$ & $\| \tau \|^2$ & -0.0001 \\
        $r_{\dot{q}} $ & $\| \dot{q} \|^2$ & -0.0001 \\
        $r_{\ddot{q}}$ & $\| \ddot{q} \|^2$ & -0.0001 \\
        \midrule
        4-legged locomotion & & \\ 
        $r_{lin~vel}$ & $e^{\|\dot{x}_{target,~xy} - \dot{x}\|^2 / 0.25}$ & 1.5 \\
        $r_{ang~vel}$ & $e^{\|\omega_{target,~z} - \omega\|^2 / 0.25}$ & 1.5 \\
        \midrule
        Ducking & & \\
        $r_{duck}$ & $ e^{0.8*|x_{goal}-x|}$ & 2 \\
        \midrule
        Stand-up & & \\
        see Tab. \ref{table:standup_reward} & & \\
         \bottomrule
    \end{tabular}
    \end{center}
\end{table}

\begin{table}
    \caption{Rewards for AOW standing up, sitting down, and navigating while standing}
    \label{table:standup_reward}
    \begin{center}
    \begin{tabular}{@{}l|ll@{}}\toprule
        symbols & description \\ 
        &&\\
        $q^{robot} \in \mathbb{H}$ & Robot base-frame rotation & \\
        $p^{robot} \in \mathbb{R}^3$ & Robot base-frame position & \\
        $q$ & Joint DOF positions (excl. wheels) & \\
        $q_{hl}$ & Hind-Leg DOF position & \\
        $\alpha$ & $\angle$(robot-x axis, world z axis) \\
        $f$ & Feet on ground (binary) \\
        $s$ & Standing robots (binary) \\
        \midrule
        stand-up & formula & weight \\
        &&\\
        $r_{\alpha}$ & $\frac{\pi/2 - \alpha}{\pi/2}$  & 2 \\
        $r_{height}$ & $p_z^{robot}$  & 3 \\
        $r_{feet}$ & $f$ & -2 \\
        $r_{wheels}$ & $\sum \dot{q}_{front~wheels}^2 * (1-f)$ & -0.003 \\
        $r_{shoulder}$ & $\|q_{shoulder}\|^2$ & -1 \\
        $r_{stand~pose}$ & $exp(-0.1 * \| q_{hl} - q_{0,~hl} \|^2)$ & 1 \\
         \midrule
        sit-down & & weight \\
        &&\\
        $r_{un-stand}$ & $max(\frac{\pi/2 - \alpha}{\pi/2} * 3, 0)$ & -3 \\
        $r_{sit-down}$ & $\frac{min(\alpha, \pi/2)}{\pi/2}$ & 2.65 \\
        $r_{dof~vel}$ & $\|\dot{q}\|^2$ & -0.015 \\
        $r_{dof~pos}$ & $\exp(-0.5 * \|q_0 - q\|^2) * \frac{\alpha}{\pi/2}$ & 3 \\
         \midrule
        navigation & & weight \\
        &&\\
        $r_{track~lin}$ & $exp(- 4 * \|\dot{x}_{des} + \dot{p}_{local, z}^{robot}\|^2) * s$ & 2 \\
        $r_{track~ang}$ & $exp(- 4 * \|\omega_{des} - \omega_{local, x}^{robot}\|^2) * s$ & 2 \\
         \bottomrule
    \end{tabular}
    \end{center}
\end{table}

%% file: 4_conclusion.tex
\section{CONCLUSIONS}
\label{sec:conclusions}
This work introduces Multi-AMP, with which we automate the imitation objective and motion selection process of multiple motion priors without heuristics. Our experimental section shows that we can simultaneously learn different styles and skills in a single policy. Furthermore, our approach can intentionally switch between these styles and skills, whereby also data-free styles are supported. Various multi-style policies are successfully deployed on a wheeled-legged robot. To this end, we show different combinations of skills such as walking, ducking, standing up on the hind legs, navigating on two wheels, and sitting down on all four legs again. We avoid tedious reward function tuning by training the sit-down motions with a motion prior gained from reversing a stand-up recording. Furthermore, we note that similar performances as in the single-style case can be expected even when learning multiple styles simultaneously. We conclude that Multi-AMP and its predecessor AMP \cite{peng_ma_abbeel_levine_kanazawa_2021} are promising steps towards a possible future without style-reward function tuning in \ac{RL}. However, even though less time is invested in tuning reward functions, more time is required to generate motion priors, which is in most cases not available for specific tasks.

To the best of our knowledge, this is the first time that a quadruped-humanoid transformation is shown on a real robot, challenging how we categorize multi-legged robots. Over the next few years, this skill will further expand the possibilities of wheeled quadrupeds by opening doors, grabbing packages, and many more use-cases.